\documentclass{article}

\usepackage{graphicx}
\usepackage{subcaption}
\usepackage[numbers]{natbib}
\usepackage{amsmath}
\usepackage{sidecap}

 \usepackage[preprint]{nips_2018}

\usepackage[utf8]{inputenc} % allow utf-8 input
\usepackage[T1]{fontenc}    % use 8-bit T1 fonts
\usepackage{hyperref}       % hyperlinks
\usepackage{url}            % simple URL typesetting
\usepackage{booktabs}       % professional-quality tables
\usepackage{amsfonts}       % blackboard math symbols
\usepackage{nicefrac}       % compact symbols for 1/2, etc.
\usepackage{microtype}      % microtypography

\title{Continual Classification Learning Using Generative Models}

\author{
  Frantzeska Lavda\thanks{University of Geneva $\&$ Geneva School of Business Administration, HES-SO} \\
  \texttt{Frantzeska.Lavda@etu.unige.ch} \\
   \And
   Jason Ramapuram$^*$ \\
   \texttt{Jason.Ramapuram@etu.unige.ch} \\
   \AND
   Magda Gregorova$^*$\\
   \texttt{Magda.Gregorova@hesge.ch} \\
  \And
   Alexandros Kalousis$^*$ \\
   \texttt{Alexandros.Kalousis@hesge.ch} \\
}

\begin{document}

\maketitle
\vspace{-4mm}
\begin{abstract}
\vspace{-3mm}
Continual learning is the ability to sequentially learn over time by accommodating knowledge while retaining previously learned experiences. Neural networks can learn multiple tasks when trained on them jointly, but cannot maintain performance on previously learned tasks when tasks are presented one at a time. This problem is called catastrophic forgetting. In this work, we propose a classification model that learns continuously from sequentially observed tasks, while preventing catastrophic forgetting. We build on the lifelong generative capabilities of \cite{jason} and extend it to the classification setting by deriving a new variational bound on the joint log-likelihood, $\log p(x, y)$.% The method is able to learn and preserve all the classification tasks it encounters without the need to store the past data or the previously learned models.
\end{abstract}

\vspace{-5mm}
\section{Introduction}
\vspace{-2.5mm}

\label{sec:intro}
%\vspace{-1mm}

%story
Continual learning tries to mimic the ability of humans to retain or accumulate previous knowledge and use it to solve future problems with possible adaptations.
In this paper we propose a new method for continual learning in the classification setting. %The network receives a sequence of classification tasks. The semantics of the classification remain the same across the tasks, but each is generated from a different distribution. During training the data of each task are seen just once. The network does not have access to the old data, it only has access to data from the current task.
Our model combines the encoder and decoder of a variational autoencoder (VAE) \citep{kingma2013auto} with a classifier. To do this we derive a new variational bound on the joint log-likelihood $\log p(x,y)$.

To enable the continual discriminative learning we build on the work of Ramapuram et al. \cite{jason} on lifelong generative modelling. The model has a student-teacher architecture (similar to that in distillation methods, \cite{hinton2015distilling}, \cite{furlanello2016active}), where the teacher contains a summary of all past distributions and is able to generate data from the previous tasks once we no longer have access to the original data.
Every time a new task arrives, a student is trained on the new data together with the data generated by the teacher from the old tasks. The proposed method thus does not need to store the previous models (it only stores their summary within the teacher model) nor data from the previous tasks (it can generate them using the teacher model).

\vspace{-2.5mm}

\subsection{Related work}
\vspace{-2.5mm}

\label{subsec:ralated}

Several approaches have been proposed to solve catastrophic forgetting over the last few years. We can roughly distinguish 2 streams of work: a) methods that rely on a dynamic architecture that evolves as they see new tasks b)methods with regularization approaches that constrain the models learned in new tasks so that the network avoids modifying the important parameters of the previous tasks. In dynamic architectures parameters of the models learned on the old tasks are passed over to the new tasks while the past models for each task are preserved (\cite{rusu2016progressive}, \cite{fernando2017pathnet}).  In contrast, our method does not need to keep the past models. Regularization approaches (\cite{ewc}, \cite{si}) impose constrains to the objective function to minimize changes in parameters important for previous tasks. However, these methods need to store the parameters of the previous tasks, something that is not required in our proposed method. 

In Variational Continual Learning, \citep{vcl}, the authors propose a method which is applicable to discriminative and generative models but not both at the same time while our method is. While VCL shows rather impressive results, it achieves those relying on the reuse of some of the previous data through the use of \emph{core-sets} and by maintaining task-specific parameters, called \emph{head networks}. It therefore relaxes the continual learning paradigms of no access to past data and no storage of past task-specific models; paradigms that our method fully takes on board.

\vspace{-3mm}

\section{Model}
\vspace{-2.5mm}

\label{sec:model}
%Following the model proposed by \cite{jason} we mitigate the catastrophic forgetting problem in classification settings. We implement that by deriving a new variational bound on the joint log-likelihhod, $\log p(x,y) =  \log p(x)p(y \vert x) = \log p(x) + \log p(y \vert x)$, and maximizing this instead of the original one. 
In the continual classification setting, we deal with data that come sequentially in pairs $(\mathbf{X, Y}) = \{ (x_1, y_1), ... , (x_n, y_n)\}$. For each task $j$ the network receives a new data set $\{(x_j , y_j)\}$ and does not have access to any of the data sets of the previously seen tasks. %While the data sets of individual tasks and their generative distributions are different, the semantics of the classification remain the same for all the tasks. 

\begin{figure}[h]
\vspace{-5mm}
\centering
\includegraphics[width=0.22\linewidth]{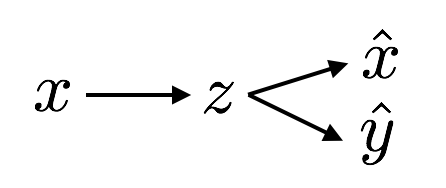}
\caption[Graphical model]{Graphical model}
\label{fig:graphical_model}
\end{figure}

To perform the classification, we use a latent variable model as shown in Fig.\ref{fig:graphical_model}. In this model, each observation $x$ has a corresponding latent variable $z$, that is used to generate the correct label class $y$. The joint distribution of the latent variable model that we consider factorizes as $p(x,y,z) =  p(x \vert z) p(y \vert z) p(z)$ where $(x,y)$ are labeled data pairs and $z$ are the latent variables. The data variables $x, y$ are assumed to be conditionally independent given the latent variables $z$, $((x \perp y) \vert z)$, such that $p(x,y \vert z) = p(x \vert z) p(y \vert z)$. 
\vspace{-1mm}
\vspace{-4mm}

Following the classical VAE approach we will use variational inference to approximate the intractable posterior $p(z \vert x,y)$.  Instead of the natural $q(z\vert x,y)$ we use $q_{\phi}(z \vert x)$ to approximate the true posterior $p(z \vert x, y)$ since in the test phase of the classification $y$ is not available. To measure the similarity between the true posterior $p(z|x,y)$ and its approximation $q(z|x)$ we minimize the Kullback-Leibler divergence between the approximate posterior and the true posterior.
\begin{equation}
D_{KL}\left( q_{\phi}(z \vert x) \Vert p(z \vert x, y) \right) =  -E_{q_{\phi}(z \vert x)} \left[ \log p(x, y, z) - \log q_{\phi}(z \vert x) \right]+  \log p(x, y) 
\label{eq:KL}
\end{equation}
The term $\log p(x, y)$ in Eq.\ref{eq:KL} is a constant. This means that in order to minimize the KL-divergence we minimize  $-E_{q(z \vert x)} \left[ \log p(x, y, z) - \log q_{\phi}(z \vert x) \ \right] = -L(x, y)$ which is equivalent to maximizing $L(x, y)$.
\begin{equation}
L(x, y) = E_{q_{\phi}(z \vert x)} \left[ \log p(x, y, z) - \log q_{\phi}(z \vert x) \ \right]
\label{eq:loss_joint}
\end{equation}
Rearranging Eq.\ref{eq:KL} as:
\begin{align}
%\begin{array}{r@{}l@{\qquad}l}
\log p(x, y) &= E_{q_{\phi}(z \vert x)} \left[ \log p(x, y, z) - \log q_{\phi}(z \vert x) \ \right] + D_{KL}\left( q_{\phi}(z \vert x) \Vert p(z \vert x, y) \right) \nonumber\\ %[\jot]
&= L(x, y) + D_{KL}\left( q_{\phi}(z \vert x) \Vert p(z \vert x, y) \right)
%\end{array}
\label{eq:fran_joint_full}
\end{align}
we can see that the $L(x, y)$ is a lower bound of the  \emph{joint} log-likelihood, $\log p(x, y)$: a new variational bound for the joint generative and discriminative VAE learning. 
%When the fit is not perfect, given that the KL-divergence is always positive, we have  $ \log p(x, y) > L(x, y)$.

To gain better intuition into our newly derived variational bound, we show the relation to the classical ELBO (variational bound on the marginal likelihood $p(x)$) used in VAEs. Rearranging the terms in Eq.\ref{eq:fran_joint_full},  under the conditional independence assumption $p(x,y|z) = p(x|z)p(y|z)$ and using the fact that the KL-divergence is always positive, we arrive at:
%\begin{equation}
%\begin{array}{r@{}l@{\qquad}l}
%\log p(x, y) &\geq L(x, y) = E_{q_{\phi}(z \vert x)} \left[ \log p(x, y, z) - \log q_{\phi}(z \vert x)  \right] \nonumber\\ [\jot]
%&=  \underbrace{E_{q_{\phi}(z \vert x)} \left[ \log p(x \vert z)\right] - D_{KL}\left( q_{\phi}(z \vert x) \Vert p(z) \right)}_{\text{ELBO}} + \underbrace{E_{q_{\phi}(z \vert x)} \left[ \log p(y \vert z)\right]}_{\text{classification \ loss}}
%\label{eq:boundFRAN}
%\end{array}
%\end{equation}
\begin{align}
%\begin{array}{r@{}l@{\qquad}l}
\log p(x, y) &\geq L(x, y) = E_{q_{\phi}(z \vert x)} \left[ \log p(x, y, z) - \log q_{\phi}(z \vert x)  \right] \nonumber\\ %[\jot]
&=  \underbrace{E_{q_{\phi}(z \vert x)} \left[ \log p(x \vert z)\right] - D_{KL}\left( q_{\phi}(z \vert x) \Vert p(z) \right)}_{\text{ELBO}} + \underbrace{E_{q_{\phi}(z \vert x)} \left[ \log p(y \vert z)\right]}_{\text{classification \ loss}}
\label{eq:boundFRAN}
%\end{array}
\end{align}
The first term $ E_{q(z \vert x)}\left[\log p(x \vert  z)\right] - D_{KL}(q(z \vert x) \Vert p(z)) $, as in the standard VAE  is the variational bound on the marginal likelihood $p(x)$ (ELBO). %$ E_{q(z \vert x)}\left[\log p(x \vert  z)\right]$ is the expectation of the conditional log-likelihood of the input $x$ on the latent variable $z$, the reconstruction loss, and $D_{KL}(q(z \vert x) \Vert p(z))$ is the Kullback-Leibler divergence between the prior distribution $p(z)$ and the learned latent posterior $q(z \vert x)$. 
The second term $E_{q(z \vert x)}\left[\log p(y \vert  z) \right]$ is the expectation of the conditional log-likelihood of the labels $y$ on the latent variable $z$, the classification loss. This term allows our variational bound to be used in classification settings. 
This means that we solved the two problems of producing the labels $y$, and  generating input data $x$ jointly, resulting in a common latent variable $z$ which is good for classification and reconstruction at the same time.

%%%%%%%%%% MG %%%%%%%%%%%%%% REPLACE THIS

%%So how about this story about the proportions equal to 1 you told us last time ... kind of why your approximations is valid?
%An alternative way to argue about the validity of our KL divergence in Eq.\eqref{eq:KL} is that under the conditional independent assumption $\frac{p(z \vert x, y)}{p(z \vert x)} = \frac{p(y \vert z}{p(y \vert x)}$. Assuming that the $z$ summarizes $x$, we can use $z$ to classify $y$ instead of $x$. This means that $\frac{p(y \vert z}{p(y \vert x)}=1 \Rightarrow \frac{p(z \vert x, y)}{p(z \vert x)}=1$. Approximating the the intractable posterior $p(z \vert x)$ with $q_{\phi}(z \vert x)$ results in $\frac{p(z \vert x, y)}{q_{\phi}(z \vert x)}\approx 1$. The latter means that we can approximate the true posterior $p(z|x,y)$ with the approximation $q(z|x)$.  %\textbf{The latter means that the KL divergence between the true posterior $p(z|x,y)$ and the approximation $q(z|x)$ is almost equal to zero validating the fact that we can approximate $p(z|x,y)$ with $q(z|x)$ OR The latter means that we can approximate the true posterior $p(z|x,y)$ with the approximation $q(z|x)$ }.

%%%%%%%%%% MG %%%%%%%%%%%%%% WITH THIS

Furthermore, it is easy to show that under our conditional independence assumption $p(z \vert x, y) / p(z \vert x) = p(y \vert z) / p(y \vert x)$.
Assuming that $z$ summarizes $x$ well for the classification of $y$ ($p(y \vert z) \approx p(y \vert x)$) both of the ratios are close to 1. 
% we have the ratios $p(y \vert z) / p(y \vert x) =  p(z \vert x, y) / p(z \vert x) = 1$.
Replacing the intractable posterior $p(z \vert x)$ by the approximation $q_{\phi}(z \vert x)$ results in $p(z \vert x, y) / q_{\phi}(z \vert x) \approx 1$ which is what the minimization of the KL-divergence in Eq.\eqref{eq:KL} tries to achieve. 
This therefore provides and alternative argument for the validity of our approach described above.

%%%%%%%%%% MG %%%%%%%%%%%%%% END

Our goal in this paper is to correctly classify data from different tasks that arrive continuously, requiring us to handle the catastrophic forgetting problem. For this we use the lifelong generative ability of \cite{jason} and extend their VAE based generative model to include a classifier that remembers all the classification tasks it has seen before. The method uses a dual architecture based on a student-teacher model. The main goal of the student model is to classify the input data. The teacher model's role is to preserve the memory of the previously learned tasks and to pass this knowledge onto the student. 

Both the teacher and the student consist of an encoder $q_{\phi^m}(z \vert x)$, a decoder $p_{\theta_x^m}(x \vert z)$ and newly a classifier $p_{\theta_y^m}(y \vert z)$ following the graphical model in Fig.\ref{fig:graphical_model}. In the above notation $m \equiv t, s$ represents the teacher and student model respectively. The teacher model remembers the old tasks and generates data from them $\{(\tilde{x},\tilde{y})\}$ for the student to use in learning once the old data are no longer available. The student model learns to generate and classify over the new labeled data pairs $\{(x,y)\}$ and the old-task data generted by the teacher $\{(\tilde{x},\tilde{y})\}$. Every time a new task is initiated, the student passes the latest parameters to the teacher and starts learning over data from the new task augmented with data generated by the teacher from all the previous tasks. In this way the acquired information of the previous tasks is preserved and the proposed model learns to classify correctly even over data distributions seen in previous tasks. The proposed architecture does not need to store the task-specific models for the previous data distributions nor the previous data themselves.
\begin{figure}[h]
\centering
\begin{subfigure}[b]{0.445\textwidth}
   \includegraphics[width=0.89\linewidth]{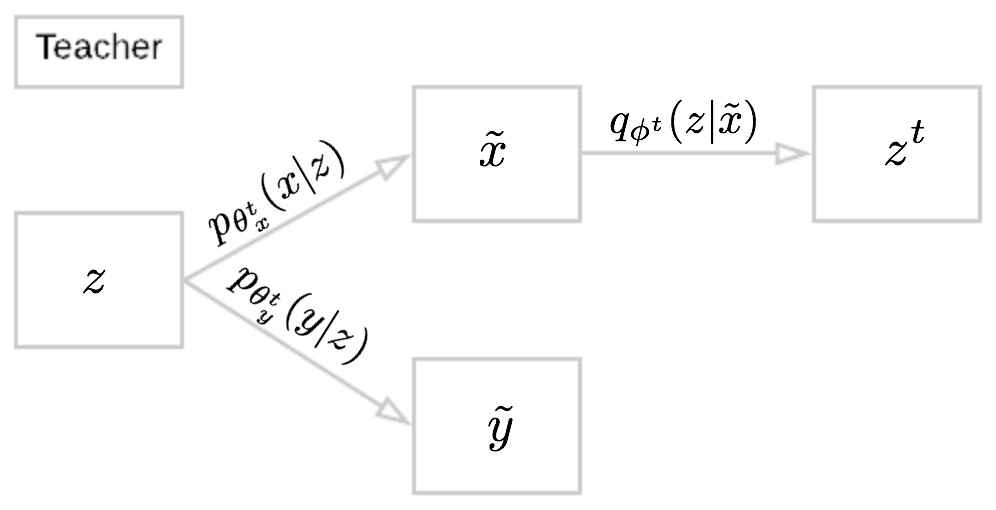}
   \caption{}
   \label{fig:teacher} 
\end{subfigure}
\begin{subfigure}[b]{0.445\textwidth}
   \includegraphics[width=0.89\linewidth]{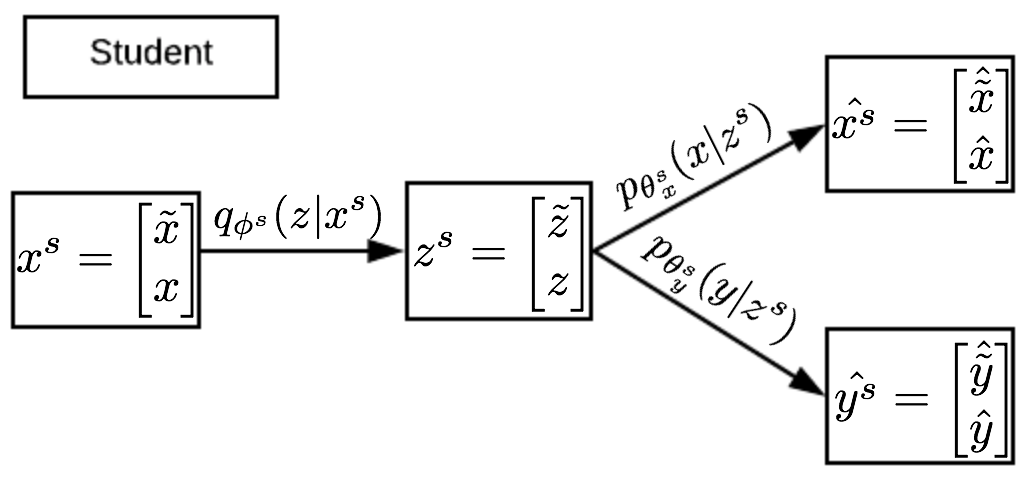}
   \caption{}
   \label{fig:student}
\end{subfigure}
\caption[Student Teacher model]{The architecture of the learning procedure. Fig.\ref{fig:teacher} The teacher model generates input-output pairs from the previously seen tasks and passes them onto the student. Moreover the teacher evaluates the posterior $ q(z \vert \tilde{x})$ Fig.\ref{fig:student} The student model learns to classify and generate new data augmented by data from the teacher.}
\label{fig:s-t}
\end{figure}
\vspace{-3mm}

%The learning process of the student model can be briefly described as follows: for a given observation $x^s$ learn a) the approximate posterior $q_{\phi^s}(z \vert x^s)$ b) the distribution $p_{\theta_x^s}(x \vert z^s)$ to reconstruct $\hat{x^s}$ c) the distribution $p_{\theta_y^s}(y \vert z^s)$ the predicted labels $\hat{y^s}$.
 %The main goal of the student model is to classify the input data. We achieve this through the use of the teacher model whose role is to preserve the memory of the previously learned tasks and to pass this knowledge onto the student. The teacher samples latent variables from the prior $z \sim P(z)$ uses the decoder and the classifier to produce samples $\tilde{x} \sim p_{\theta_x^t}(x \vert z)$ and $\tilde{y} \sim p_{\theta_y^t}(y \vert z)$. The generated input-output pairs $(\tilde{x}, \tilde{y})$ are then passed onto the student together with the data of a new task. %In this way we transfer the information about the past tasks. The student then learns the distributions over the new incoming data while accommodating for the knowledge obtained from the teacher. In this way the acquired information of the previous tasks is preserved and the proposed model learns to classify correctly even over data distributions seen in previous tasks. 
 %When a new task arrives the teacher is dropped and the student parameters are copied to the new teacher. The new student keep the parameters of the previous student and continuous training. 
The student optimizes the variational bound of the joint log-likelihood $\log p(x,y)$ Eq.\eqref{eq:boundFRAN} instead of the marginal log-likelihood $\log p(x)$ over which the classical VAE operates. As a result our model is able to both generate the input data $x$ and learn the labels $y$ at the same time. We should note that previous approaches to classification with VAEs (\cite{kingma2014semi}) do so by adding an ad-hoc manner to the VAE optimization function terms that relate to classification performance. Here we naturally extend the VAE setting to classification. 

 Following \cite{jason} we add an additional term ($D_{KL}[q_{\phi^s} (z^s \vert \tilde{x}) \Vert q_{\phi^t}(z^t \vert \tilde{x})$) to our objective to preserve the posterior representation of all previous tasks to speed up the training and a negative information gain regularizer, $L_I(z, \tilde{x})$, between the latent representation $z$ and the generated data $\tilde{x}$ from the teacher. The final loss that we optimize is given by Eq.\ref{eq:loss_with_consistency}.
%By maximizing the proposed variational bound (Eq.\ref{eq:boundFRAN}) on the joint log-likelihood, $(\log p(x,y) =  \log p(x)p(y \vert x) = \log p(x) + \log p(y \vert x))$, instead of the original marginal bound we can see the student as two problems that are optimized jointly. In the first one we have to generate the input data $x$, and in the second one we have to produce the labels $y$. Therefore, our model learns continuously different task in the classification setting and at the same time, as a byproduct of our bound, it also reconstructs the input. Following \cite{jason} we add an additional term ($D_{KL}[q_{\phi^s} (z^s \vert \tilde{x}) \Vert q_{\phi^t}(z^t \vert \tilde{x})$) to our objective ti preserve the posterior representation of all previous tasks to speed up the training. The final loss that we optimize is given by Eq.\ref{eq:loss_with_consistency}.
\vspace{-1mm}
\begin{small}
\begin{equation}
E_{q_{\phi^s}(z \vert x^s)}\left[\log p_{\theta_x^s}(x \vert  z^s) + \log p_{\theta_y^s}(y \vert  z^s)\right] -D_{KL}(q(z \vert x) \Vert p(z)) -  D_{KL}[q_{\phi^s} (z^s \vert \tilde{x}) \Vert q_{\phi^t}(z^t \vert \tilde{x})] - L_I(z, \tilde{x})
\label{eq:loss_with_consistency}
\end{equation}
\end{small}

\vspace{-7mm}
\section{Experiments}
\vspace{-2.5mm}

In this section we present preliminary results achieved with the proposed model. We investigate the problem of whether our model is able to learn a set of different tasks that are coming in sequence without forgetting the previously trained tasks.

We evaluated our approach for continual learning on permuted MNIST \cite{lecunmnist}, \cite{goodfellow2013empirical}. 
Each task is a 10-way classification (0-9 digits) over images with the pixels shuffled by a random fixed permutation.
We train on a sequence of 5 tasks (original MNIST and 4 random permutations).
After the training of each task we allow no further training or access to that task's data set.
For training we process the data in mini batches of 256 (random data shuffling) and use early stopping on the classification accuracy.

We use two baseline models for comparisons.
The first is a standard VAE augmented by our classifier (vae-cl) using our variational bound but without the teacher-student architecture.
In the second, we adapt the elastic weight consolidation (EWC) regularisation approach of \cite{ewc} to our setting.
We use the teacher here to keep the summary of all the previous distributions\footnote{In our EWC baseline the teacher is not used to generate data for the student} and employ the EWC-like regularisation $\sum_i F_i (\psi_i^s - \psi_i^t)^2$ over the parameters of the teacher and student $(\psi^m = [\theta^m, \phi^m])$ models where $F_i = \text{diag} E[(\partial L_{\psi}(x, y) / \partial \psi)^2] $.%($F_i$ are the diagonal elements of the fisher information matrix and the fisher is calculated using our variational bound, Eq.\eqref{eq:boundFRAN}).

\begin{figure}[h]
\centering
\begin{subfigure}[b]{0.449\textwidth}
   \includegraphics[width=0.95\linewidth]{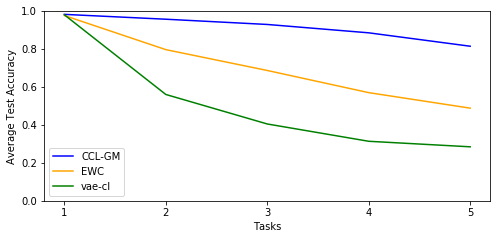}
   \caption{Average test classification accuracy}
   \label{fig:average_accuracy} 
\end{subfigure}
\begin{subfigure}[b]{0.449\textwidth}
   \includegraphics[width=0.95\linewidth]{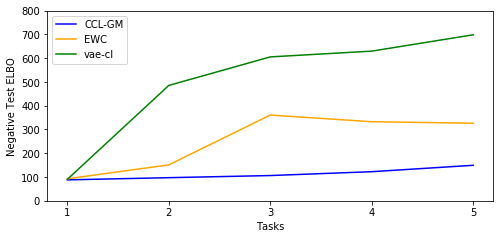}
   \caption{Average test negative reconstruction ELBO}
   \label{fig:averege_elbo}
\end{subfigure}
\caption{Average performance over all learned tasks from the permuted MNIST data set as a function of the number of tasks. Our approach, CCL-GM maintains high accuracy and low negative ELBO as the number of tasks increases. Vanilia VAE our classifier performs far worse. EWC  degrades less severely, but still forgets the previous tasks}\label{fig:exp1}
\end{figure}

We measure the performance by the ability of the network to solve all tasks seen up till the current point.
For all tested methods we performed a random hyper-parameter sweep over convolutional and dense network architectures.
We present the results of the best obtained models\footnote{Convolutional for ours and vae-cl, dense for EWC} in Fig. \ref{fig:exp1}.
For the naive vae-cl method the performance drops dramatically already when the training regime switches from the MNIST to the first permuted task. 
For the EWC method the performance after the first task degrades less severely, but it still forgets the previous tasks.
Our model, continual classification learning using generative models (CCL-GM) retains high average classification accuracy Fig. \ref{fig:average_accuracy} and low average reconstruction ELBO \ref{fig:averege_elbo}. This shows that our model is able to learn continuously and concurrently for both classification and generation.

%%%%%%%%%% MG %%%%%%%%%%%%%% END

%\vspace{-2mm}

%%%%%%%%%% MG %%%%%%%%%%%%%% REPLACE THIS

%Repeating the same experiment using now a sequence of three different tasks, MNIST, FashionMNIST,
%permuted MNIST we can verify that our model mitigates the catastrophic forgetting and it outperforms
%the two baselines, as it is demonstrated in Fig.\ref{fig:MFP}

%%%%%%%%%% MG %%%%%%%%%%%%%% WITH THIS

\begin{figure}[h]
\centering
\begin{subfigure}[b]{0.449\textwidth}
   \includegraphics[width=0.95\linewidth]{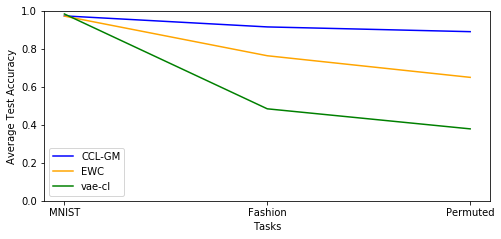}
   \caption{Average test classification accuracy}
   \label{fig:MFP_average_accuracy} 
\end{subfigure}
\begin{subfigure}[b]{0.449\textwidth}
   \includegraphics[width=0.95\linewidth]{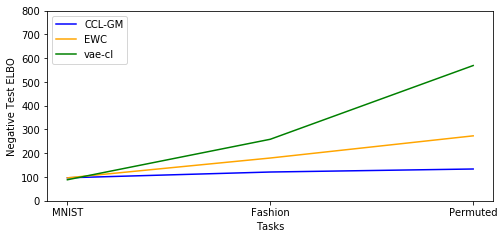}
   \caption{Average test negative reconstruction ELBO}
   \label{fig:MFP_averege_elbo}
\end{subfigure}
\caption{Average performance over all learned tasks. }
\label{fig:MFP}
\end{figure}

To support our initial results from the above experiments, we conducted a second set of experiments on a sequence of three different tasks: MNIST, FashionMNIST \cite{xiao2017fashion} and one MNIST permutation. The results presented in Fig. \ref{fig:MFP} show that our method outperforms the baselines and confirm our preliminary conclusions that our new model CCL-GM has the ability to mitigate catastrophic forgetting in joint generative and discriminative problems.

%%%%%%%%%% MG %%%%%%%%%%%%%% END

\vspace{-5mm}
\section{Conclusion}
\vspace{-2.5mm}

\label{sec:conclusion}
In this work we propose a method to address continual learning in the classification setting. %Our network receives data from different distributions $P(X,Y)$.
We use a generative model to generate input-output pairs from the previously learned tasks and use these to augment the data of the current tasks for further training.  
In this way our classification model overcomes catastrophic forgetting. Our model does not reuse data nor previous task-specific models and it continuously learns to concurrently classify and reconstruct data over a number of different tasks.

\newpage
\bibliographystyle{plain}
\bibliography{bib-icml}

\end{document}